# ArabJobs: A Multinational Corpus of Arabic Job Ads

Mo El-Haj
VinUniversity, Vietnam
Lancaster University, UK
`elhaj.m@vinuni.edu.vn`
`m.el-haj@lancaster.ac.uk`

## Abstract

ArabJobs is a publicly available corpus of Arabic job advertisements collected from Egypt, Jordan, Saudi Arabia, and the United Arab Emirates. Comprising over 8,500 postings and more than 550,000 words, the dataset captures linguistic, regional, and socio-economic variation in the Arab labour market. We present analyses of gender representation and occupational structure, and highlight dialectal variation across ads, which offers opportunities for future research. We also demonstrate applications such as salary estimation and job category normalisation using large language models, alongside benchmark tasks for gender bias detection and profession classification. The findings show the utility of ArabJobs for fairness-aware Arabic NLP and labour market research. The dataset is publicly available on GitHub: https://github.com/drelhaj/ArabJobs.

## 1 Introduction

The expansion of Arabic Natural Language Processing (NLP) research has supported progress in areas such as sentiment analysis, named entity recognition, and machine translation (Antoun et al., 2020). However, the field continues to face a shortage of datasets that are both linguistically diverse and representative of socio-economic realities. Job advertisements offer a valuable lens into labour market discourse, often encoding assumptions about gender roles, social hierarchies, and regional language practices. Prior research has demonstrated the presence of gender bias in such texts and stressed the importance of computational techniques to detect and reduce these biases (Dikshit et al., 2024a).

Despite the importance of employment-related text for sociolinguistic and fairness-oriented NLP, no publicly available Arabic corpus exists that captures the structure and linguistic diversity of job advertisements across multiple Arab countries. To our knowledge, no prior datasets have been released in this domain, and existing work on Arabic job-related text is either non-existent or inaccessible. To address this gap, we present **ArabJobs**, a corpus of Arabic job advertisements collected from four countries—Egypt, Jordan, the United Arab Emirates, and Saudi Arabia. The dataset includes structured fields such as job title, location, and salary, as well as unstructured job descriptions, offering broad coverage across sectors and dialects.

## 2 Literature Review

Despite recent advances in Arabic NLP, the field continues to face a shortage of domain-specific and socio-linguistically diverse corpora. While general-purpose datasets and language models have been developed for Arabic (Antoun et al., 2020; El-Haj, 2020; Alhafni et al., 2024; Daoud et al., 2025; El-Haj and Ezzini, 2024; Elmadani et al., 2025; El-Haj et al., 2024), resources grounded in real-world contexts—such as employment, health, or finance—remain rare. This limits the development of systems capable of modelling Arabic in ways that reflect regional variation, social practices, and occupational language.

For English, job advertisement datasets have enabled a range of impactful studies, particularly in the analysis of bias, fairness, and labour market discourse. For example, recruitment corpora have been used to reveal implicit gender stereotypes in job descriptions (Dikshit et al., 2024b), providing empirical foundations for bias detection tools and fairness-aware text generation. Such work has underscored the value of job ads as a lens into both linguis-tic and socio-economic structures. However, no comparable resource exists for Arabic, leaving a significant gap in our ability to conduct similar analyses across the Arab region. The ArabJobs corpus fills this gap by introducing the first publicly available, multi-country corpus of Arabic job advertisements. Covering posts from Egypt, Jordan, Saudi Arabia, and the UAE, it enables the study of regional dialect use, gender representation, and occupational framing in real-world labour discourse. The corpus is designed to support downstream NLP tasks and facilitate investigations into sociolinguistic variation in a structured, professionally relevant setting.

Prior work on gender and dialect in Arabic NLP further highlights the importance of such domain-grounded corpora (Alhafni et al., 2022). Bias detection and mitigation strategies have largely been confined to general-purpose or translated datasets, with limited exploration of high-stakes, real-world domains like employment. Tools such as AraWEAT (Lauscher et al., 2020) and the Arabic Parallel Gender Corpus (Alhafni et al., 2022) provide important foundations for modelling gender sensitivity, while dialect classification benchmarks like MADAR (Bouamor et al., 2019), NADI (Abdul-Mageed et al., 2020), and ALDi (Keleg et al., 2023) offer frameworks for analysing linguistic variation. Yet, these efforts often operate independently of professional or institutional contexts. By anchoring linguistic analysis in the domain of job advertising—where language directly impacts access to opportunity—the ArabJobs corpus offers a new lens for examining structural inequality, dialectal salience, and cultural norms embedded in Arabic textual data. Our study explores how gendered language and job category structures manifest in Arabic job advertisements. We also extend research directions commonly pursued in English NLP, such as implicit gender bias detection and the use of LLMs for salary estimation and job classification—demonstrating how a domain-specific corpus can support analogous investigations in Arabic and open new avenues for NLP research in the region.

## 3 ArabJobs Corpus

The ArabJobs corpus is the first large-scale, publicly available dataset of Arabic job advertisements, supporting research in NLP, labour market analysis, sociolinguistics, and computational social science. It contains **8,546** ads totalling over **550,000 words**, collected from Egypt, Jordan, Saudi Arabia, and the UAE. These cover a wide range of sectors and reflect regional linguistic and socio-economic variation.

Each entry includes structured fields such as job title, location, salary (or estimate), gender preference, and free-text descriptions. Table 1 presents a breakdown by country, showing the number of ads, gender targeting (male, female, or neutral), and average word count per post. This dataset enables nuanced analyses of how job markets communicate expectations and supports investigations into gendered language, occupational framing, and fairness in employment discourse.

| Country      | Ads   | Male  | Female | Neutral | Avg. Word Count |
|--------------|-------|-------|--------|---------|-----------------|
| Egypt        | 3,598 | 2,085 | 313    | 1,200   | 58.88           |
| Jordan       | 1,147 | 498   | 370    | 279     | 47.49           |
| Saudi Arabia | 1,854 | 972   | 264    | 618     | 116.65          |
| UAE          | 1,947 | 1,212 | 427    | 308     | 28.57           |

Table 1: Job Advertisement Statistics by Country

As shown in Figure 1, Egypt and the UAE account for the largest number of job advertisements in the corpus, followed by Saudi Arabia and Jordan. These differences likely reflect underlying labour market dynamics and platform usage across the region. The breakdown also reveals notable variation in posting volume and length, both of which are relevant for downstream analyses of language use and content structure.

### 3.1 Data Collection

The **ArabJobs** corpus was constructed by scraping Arabic job advertisements from seven publicly accessible recruitment platforms across the MENA region. We complied with all `robots.txt` restrictions, excluded paywalled or login-protected content, and implemented rate limiting to ensure respectful data collection. All personally identifiable information—such as names, emails, and phone numbers—was removed during post-processing (see Section 8 for further details).

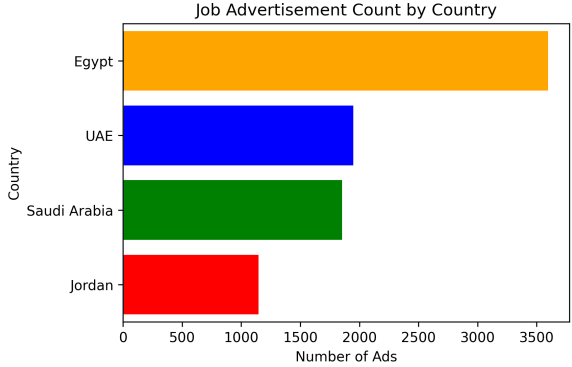

Figure 1: Distribution of job advertisements across four countries in the ArabJobs corpus.

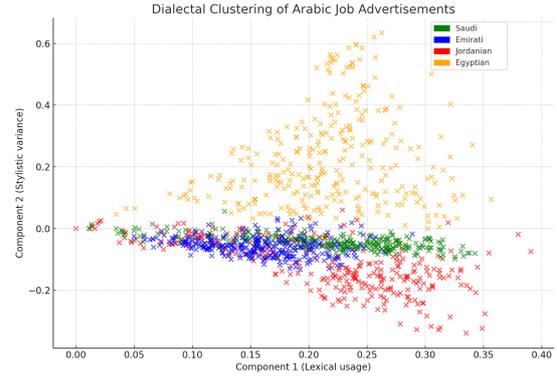

Figure 2: Dialectal clustering. Component 1 captures lexical variation; Component 2 reflects stylistic differences.

Each job entry in the corpus includes structured fields such as `job_title`, `location`, `salary`, `gender`, `description`, and `country`. Additionally, the dataset contains fields generated via LLM-based inference—`profession`, `salary_local`, `salary_usd`, `job_category`, and `sub_category`—which were subsequently verified by native Arabic-speaking annotators.

## 4 Dialectal Variation and Code-Switching Analysis

Although ArabJobs does not explicitly annotate dialects, its multinational scope naturally captures regional linguistic variation. To explore this, we conducted an unsupervised analysis using job descriptions from Egypt, Jordan, Saudi Arabia, and the UAE.

We sampled 1,500 ads per country to ensure a balanced dataset and represented job descriptions using TF-IDF features. Dimensionality reduction via Truncated Singular Value Decomposition (SVD) revealed clear regional clusters (Figure 2). Saudi and Emirati ads (Gulf dialects) clustered closely, while Egyptian and Jordanian postings formed separate regions, reflecting variation in dialect and register. For instance, Jordanian ads for female beauty salons often use صالون سيدات, whereas terms preferred in Gulf ads include صالون حريمي and صالون نسائي. Dialectal differences also appear in barbering roles (حلاق, مصفف شعر, and كوافير), as well as in transport-related terms such as درايفر, عجلة, دراجة هوائية, سكوتر, ليسن, رخصة, سواقة, and رخصة قيادة.

We also analysed code-switching—English word usage within Arabic descriptions. As shown in Figure 3, ads from Jordan, Egypt, and Saudi Arabia featured more English terms (e.g., "Sales Executive", "Supervisor"), especially in sales and admin roles. In contrast, UAE postings more consistently used Arabic or Arabised terms such as 'البريد الإلكتروني', 'ايميل', 'السيرة الذاتية', and 'السي في'.

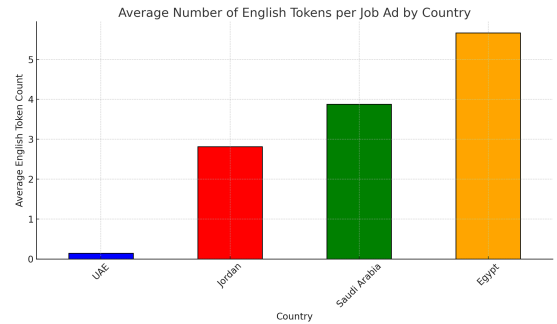

Figure 3: ArabJobs: Arabic-English Code Switching.

While job advertisements are typically composed in Modern Standard Arabic (MSA), dialectal features frequently appear, often unintentionally, even in contexts considered formal. This reflects the broader phenomenon of diglossia in Arabic, where speakers regularly shift between MSA and regional varieties. For instance, Egyptian ads may include everyday colloquialisms such as عجلة ("bike"), while Jordanian postings might favour Arabised English borrowings like سكوتر ("scooter"). These variations do not necessarily index prestige or class, but rather highlight the influence of local linguistic norms and digital writing practices. Similarly, ads for the beauty sector in Jordan may adopt familiar, community-oriented phrasing, whereas Gulf postings lean towards more formal or gender-specific expressions. These linguistic patterns reflect how Arabic speakers naturally draw from their dialects—even in written form—using them to convey relatability, express culturally grounded meanings, or enhance the communicative effectiveness of the advertisement.

## 5 Corpus Processing and Normalisation

To enable structured analysis and downstream NLP tasks, we applied several post-processing steps to enrich the raw job advertisements with additional metadata. This included inferring missing salary information, normalising inconsistent job categories, and generating standardised labels. These steps combined rule-based procedures, large language model prompting, and manual verification to improve the corpus's analytical utility.

### 5.1 Salary Estimation

The `salary` field records the original salary information when provided, either as a single figure (e.g., 3000 Emirati Dirham) or a range (e.g., 9000–11100 Egyptian Pound). However, only 3,265 job advertisements included this information. To address the substantial number of missing values, we used GPT-4 (OpenAI, 2023) to estimate salaries based on other job attributes. Rather than using the model interactively via a conversational interface, we adopted a prompt-based inference approach. Specifically, we constructed structured prompts that included 3 in-context examples drawn from the 3,265 salary-annotated ads, followed by a new instance requiring prediction (See Appendix **A**). Examples were reused across prompts, sampled by country and job category to maintain relevance. Our aim was not to introduce a novel estimation method, but to show that the dataset is structured and unambiguous enough to support downstream tasks with state-of-the-art LLMs.

To evaluate the model's predictive performance, we tested its output against the full set of 3,265 job ads with known salary values. As shown in Table 2, the model achieved a low mean absolute error (MAE) of 11.83 and a root mean square error (RMSE) of 14.84. Additionally, 98.5% of predictions fell within ±10% of the true salary, and 99.45% fell within ±20%. The Pearson correlation coefficient was 0.997, indicating a linear alignment in this simulated setup. These results demonstrate that the model performs reliably in structured inference tasks, with prediction quality that aligns well with the distribution of true values.

| Metric | Value |
| --- | --- |
| Number of Samples Evaluated | 3,265 |
| Mean Absolute Error (MAE) | 11.83 |
| Root Mean Square Error (RMSE) | 14.84 |
| Pearson Correlation ($r$) | 0.997 |
| Within ±10% of Actual Salary | 98.50% |
| Within ±20% of Actual Salary | 99.45% |

Table 2: Evaluation results for simulated salary estimation using GPT-4

To further validate the reliability of these estimates, we conducted a human evaluation. Two native Arabic-speaking annotators (Annotator 1 and Annotator 2), both fluent in Modern Standard Arabic—independently estimated salaries for a random sample of 500 job ads each. Both annotators had access to the full set of 3,265 salary-annotated ads, excluding the 500 samples they were asked to label. As with the model evaluation, salary ranges (e.g., 1000–2000) were reduced to their midpoints for comparison.

Inter-annotator agreement was high: 93% of estimates matched within a ±20% margin, and 89% within ±10%. GPT-4's predictions also aligned well with human judgement. Agreement between GPT-4 and Annotator 1 reached 85% within ±20% and 81% within ±10%, while alignment with Annotator 2 was slightly lower at 81% and 78%, respectively. These results, shown in Table 3, demonstrate that the model's estimates are both stable and broadly comparable in quality to human annotation.

The `salary_local` and `salary_usd` columns were generated for all 8,546 job advertisements as explained above. `salary_local` reflects the salary in the original currency of the job post (e.g., Jordanian Dinar, Saudi Riyal, Emirati Dirham, Egyptian Pound), while `salary_usd` provides the corresponding amount converted to US Dollars.[1]

---
[1]Conversion rates used: 1 JOD = 1.41 USD, 1 SAR

| Comparison | Agreement |
|---|---|
| A1 vs A2 @ ±10% | 0.89 |
| A1 vs A2 @ ±20% | 0.93 |
| GPT-4 vs A1 @ ±10% | 0.81 |
| GPT-4 vs A1 @ ±20% | 0.85 |
| GPT-4 vs A2 @ ±10% | 0.78 |
| GPT-4 vs A2 @ ±20% | 0.81 |

Table 3: Inter-annotator agreement for salary estimation(A1, A2: Annotators 1 and 2.)

## 5.2 Job Category Unification

The `job_category` field captures the functional sector of each job advertisement (e.g., Customer Service, Engineering). These labels were originally assigned by the source platforms[2] but varied significantly across sites due to inconsistent taxonomies—for example, مساعد إداري (Administrative Assistant), موظف استقبال (Receptionist), and سكرتير (Secretary) all describe similar roles but were labelled differently. First, all raw category names were aggregated to capture the full range of sectoral variation. Then, GPT-4 was used to cluster semantically similar labels—translating when needed. For example, خدمة العملاء, , and Customer Service / Call Centre were merged under خدمة عملاء (Customer Service).

To reduce fragmentation, rare or overlapping categories were merged under broader labels. For example, صيدلة, تمريض, طب, and الرعاية الصحية were unified under علوم ورعاية صحية (Healthcare). To retain granularity, the original profession labels were preserved in a separate `sub_category` column, enabling both general and detailed analyses (e.g., comparing nurses and pharmacists).

This process yielded a coherent taxonomy of Arabic job sectors. Table 4 summarises the resulting category distribution.

## 6 Gender Representation and Occupational Trends

The frequent use of gendered language in the ArabJobs corpus makes gender representation and bias a central focus of analysis. Gender is often explicitly stated—e.g., مطلوب موظفة[3]—or implied through gendered job titles and descriptions. This enables a detailed analysis of both explicit and implicit gender preferences across countries and job sectors.

| Arabic Category | English Translation | Ad Count |
|---|---|---|
| مبيعات | Sales | 1783 |
| فنيين وحرفيين | Technicians and Craftsmen | 960 |
| إدارة وسكرتارية | Admin and Secretarial | 777 |
| سياحة ومطاعم | Tourism and Restaurants | 733 |
| مالية ومحاسبة | Finance and Accounting | 579 |
| سيارات وميكانيك | Automotive and Mechanics | 460 |
| تسويق | Marketing | 447 |
| خدمة عملاء | Customer Service | 428 |
| هندسة | Engineering | 360 |
| خدمات تنظيف | Cleaning Services | 290 |
| موارد بشرية | Human Resources | 272 |
| رعاية صحية | Healthcare | 260 |
| صناعة وتجزئة | Manufacturing and Retail | 251 |
| صحة وجمال | Health and Beauty | 221 |
| إعلام وتصميم | Media and Design | 220 |
| أمن وحراسة | Security | 145 |
| سائقين وتوصيل | Drivers and Delivery | 145 |
| تعليم | Education | 108 |
| تكنولوجيا المعلومات | Information Technology | 69 |
| قانون ومحاماة | Law and Legal Services | 38 |
| Total | – | 8,546 |

Table 4: Distribution of job advertisements by unified job category

It is important to note that gender labels in the dataset are drawn directly from the original job platforms (see Table 1). Our use of the term "implicit gender" does not refer to inferred labels, but rather to gendered language that appears in job descriptions, such as جميلة ("beautiful") or لبقة ("well-mannered"). By contrast, "explicit gender" refers to ads that state a gender requirement directly, such as through the use of morphologically marked job titles or phrases like مطلوب موظفة ("female employee required").

### 6.1 Gender Label Distribution Across Countries

As shown in Figure 4, most job postings are directed at men, with far fewer targeting women or using neutral language. While this imbalance is consistent across countries, its extent varies, reflecting national labour market dynamics and cultural norms, highlighting the need to examine how gender is encoded in recruitment language.

---

= 0.27 USD, 1 AED = 0.27 USD, 1 EGP = 0.032 USD.

[2]We preserved the original categorisation in the `profession` field, as shown in Section 3.1.

[3]In Arabic, grammatical gender is marked morphologically. For instance, موظف (male employee) becomes موظفة (female employee) with the suffix ـة.

Figure 4: Gender distribution in Arabic job advertisements by country.

## 6.2 Gendered Occupational Patterns

The corpus spans a wide range of occupational diversity making it suitable for downstream NLP tasks involving profession classification, summarisation, and thematic bias detection. A closer analysis, however, reveals clear gender-based occupational segregation.

As shown in Figure 5, male-targeted job ads disproportionately reference technical, physical, and logistical professions—such as سائقين (technicians), مهندسين (engineers), فنيين (drivers), and مندوب مبيعات (sales agents). Industry-related roles such as أعمال ميكانيكا (mechanical work), إنتاج (manufacturing), and مقاولات (construction) are also dominant. These roles tend to prioritise skills related to physical labour, trade certifications, and logistics.

Figure 5: Word clouds of male-targeted job advertisements. Left: professions extracted from job titles. Right: weighted job categories (category size reflects its relative frequency across male-targeted ads.).

In contrast, the female-targeted word clouds in Figure 6 reveal a concentration in service, administrative, and care-related roles. Commonly mentioned positions include سكرتارية (secretarial work), تجميل (beauty), خدمة العملاء (customer service), مساعد إداري (administrative assistant), and موظفة استقبال (receptionist). These roles typically emphasise communication, hospitality, appearance, and interpersonal skills—reinforcing prevailing gender norms in the professional landscape.

Figure 6: Word clouds of female-targeted job advertisements. Left: professions from job titles. Right: gender-weighted job categories.

## 6.3 Gender-Based Salary Disparity

The descriptive statistics in Table 5 reveal a consistent salary gap across the dataset. While male-targeted job ads not only dominate in number and occupational variety, they also tend to offer higher average salaries compared to those aimed at women.

| Country | Gender | AvgLoc | AvgUSD | N |
|---|---|---|---|---|
| Egypt | female | 7079.29 | 226.54 | 313 |
| Egypt | male | 8080.22 | 258.57 | 2085 |
| Egypt | neutral | 8078.95 | 258.53 | 1200 |
| Jordan | female | 358.92 | 505.98 | 370 |
| Jordan | male | 412.73 | 581.95 | 498 |
| Jordan | neutral | 403.48 | 568.92 | 279 |
| Saudi Arabia | female | 4057.12 | 1095.43 | 264 |
| Saudi Arabia | male | 4356.65 | 1176.3 | 972 |
| Saudi Arabia | neutral | 4060.97 | 1096.47 | 618 |
| UAE | female | 3092.28 | 834.96 | 427 |
| UAE | male | 2641.01 | 713.08 | 1212 |
| UAE | neutral | 2998.43 | 809.61 | 308 |

Table 5: Average salary by country and gender. **AvgLoc**: Average salary in local currency; **AvgUSD**: Average salary in USD; **N**: Number of ads.

To quantify gender-based pay disparities, we compute the gender pay gap as the difference between the average salaries of male- and female-targeted job advertisements, relative to the female average:

$$\text{Pay Gap} = \frac{\text{Avg Salary}_{\text{male}} - \text{Avg Salary}_{\text{female}}}{\text{Avg Salary}_{\text{female}}} \quad (1)$$

A positive gap indicates that men are offered higher average salaries than women, while a negative value signals the reverse. As shown in Table 6, male-targeted roles have higher average pay in Egypt (14.14%), Jordan (15.01%), and Saudi Arabia (7.38%). The UAE is the exception, showing a negative gap of –14.6%, where female-targeted roles offer slightly higher salaries. This is largely due to sectoral distribution: the most common category in UAE ads is فنيين وحرفيين (Technicians and Craftsmen), comprising 18% of all postings and offering the lowest average pay—mostly targeted at men. .

| Country | M-USD | F-USD | Gap$ | Gap% |
|---|---|---|---|---|
| Egypt | 258.57 | 226.54 | 32.03 | 14.14% |
| Jordan | 581.95 | 505.98 | 75.97 | 15.01% |
| Saudi Arabia | 1176.3 | 1095.43 | 80.88 | 7.38% |
| UAE | 713.08 | 834.96 | -121.88 | -14.6% |

Table 6: Gender pay gap in average salaries by country. **M-USD**: Male average salary in USD; **F-USD**: Female average salary in USD; **Gap$**: Difference (M - F); **Gap%**: Percentage gap relative to female salary. Positive values indicate higher male pay.

## 6.4 Structural Gender Representation Across Job Categories

To investigate structural gender imbalances, we analysed the proportion of explicitly male- and female-targeted ads across job categories, excluding neutral listings. For each category, we calculated the percentage of male- and female-targeted ads, identified the dominant gender, and computed a **gender skew metric**—the absolute difference between male and female shares—to capture the degree of gender exclusivity.

Table 7 presents the results, ranked by descending gender skew. Certain fields show extreme male dominance, such as أمن وحراسة (Security) and فنيين وحرفيين (Technicians and Craftsmen), with over **96%** of postings targeting men. Others, like صناعة وتجزئة (Manufacturing and Retail) and سائقين وتوصيل (Drivers and Delivery), also display substantial male bias.

In contrast, categories like صحة وجمال (Health and Beauty) and تعليم (Education) are predominantly female-oriented, with over **70%** of postings directed at women. These patterns reflect deeply embedded gender norms around occupational roles.

The analysis shows that gender disparity is not limited to salaries—it is structurally

---
[3]These figures reflect unregulated online job postings and may not represent official labour market policies.

---

rooted in the allocation of roles. Addressing gender equity in the labour market requires tackling both pay gaps and access to opportunity.

| Arabic Category | English | All | %Male | %Female | Dominance | Skew (%) |
|---|---|---|---|---|---|---|
| سيارات وميكانيك | Automotive and Mechanics | 425 | 98.4 | 1.6 | Male | 96.8 |
| أمن وحراسة | Security | 118 | 98.3 | 1.7 | Male | 96.6 |
| سائقين وتوصيل | Drivers and Delivery | 124 | 97.6 | 2.4 | Male | 95.2 |
| فنيين وحرفيين | Technicians and Craftsmen | 869 | 96.5 | 3.5 | Male | 93.0 |
| هندسة | Engineering | 309 | 91.3 | 8.7 | Male | 82.6 |
| موارد بشرية | Human Resources | 184 | 89.1 | 10.9 | Male | 78.2 |
| صناعة وتجزئة | Manufacturing and Retail | 196 | 88.3 | 11.7 | Male | 76.6 |
| مالية ومحاسبة | Finance and Accounting | 306 | 81.4 | 18.6 | Male | 62.8 |
| سياحة ومطاعم | Tourism and Restaurants | 499 | 79.8 | 20.2 | Male | 59.6 |
| تكنولوجيا المعلومات | Information Technology | 34 | 79.4 | 20.6 | Male | 58.8 |
| تعليم | Education | 66 | 22.7 | 77.3 | Female | 54.6 |
| خدمات تنظيف | Cleaning Services | 231 | 74.9 | 25.1 | Male | 49.8 |
| مبيعات | Sales | 1082 | 74.5 | 25.5 | Male | 49.0 |
| إعلام وتصميم | Media and Design | 93 | 74.2 | 25.8 | Male | 48.4 |
| تسويق | Marketing | 304 | 74.0 | 26.0 | Male | 48.0 |
| رعاية صحية | Healthcare | 192 | 66.1 | 33.9 | Male | 32.2 |
| صحة وجمال | Health and Beauty | 206 | 36.4 | 63.6 | Female | 27.2 |
| خدمة عملاء | Customer Service | 297 | 37.0 | 63.0 | Female | 26.0 |
| إدارة وسكرتارية | Admin and Secretarial | 587 | 62.9 | 37.1 | Male | 25.8 |
| قانون ومحاماة | Law and Legal Services | 19 | 57.9 | 42.1 | Male | 15.8 |

Table 7: Gender skew across job categories, measured as the absolute difference between male and female ad proportions.

To better understand salary distribution across job categories, we visualised the average salaries for male- and female-targeted job advertisements, paying particular attention to dominant gender representation. Many professions show strong gender imbalances—for example, 98% of أمن وحراسة ads target men—so simply averaging all ads could produce misleading results. To account for this, we applied a dominance-aware adjustment strategy.

We began by computing the average salaries separately for male-targeted and female-targeted ads within each category. For each category, we identified the dominant gender based on the number of advertisements. The dominant gender's average salary was then given greater interpretive weight to minimise distortion from underrepresented groups. Figure 7 illustrates this comparison. The salary lines for men (solid) and women (dashed) vary across categories, with the grey bars showing the adjusted category-wise averages weighted by gender dominance.

The analysis reveals that high-paying fields like Engineering (هندسة) and IT (تكنولوجيا المعلومات) are predominantly male-targeted, with female ads in these sectors offering considerably lower average salaries—though such cases are few. In contrast, Education (تعليم), typically female-dominated, shows higher average pay for women, likely due to a small number of well-paid positions. Sales (مبيعات) and Customer Service (خدمة عملاء) are more gender-balanced and exhibit narrower salary gaps. Security (أمن وحراسة) and Drivers and Delivery (سائقين وتوصيل) remain male-exclusive, rendering female salary data in these fields negligible. Interestingly, sectors like Marketing (تسويق) and Health and Beauty (صحة وجمال) offer higher average pay for female-targeted roles, though male participation in these fields is limited. Overall, gender disparities persist not only in pay but also in access to lucrative professions, with many seemingly positive trends for women arising from isolated cases rather than systemic equality.

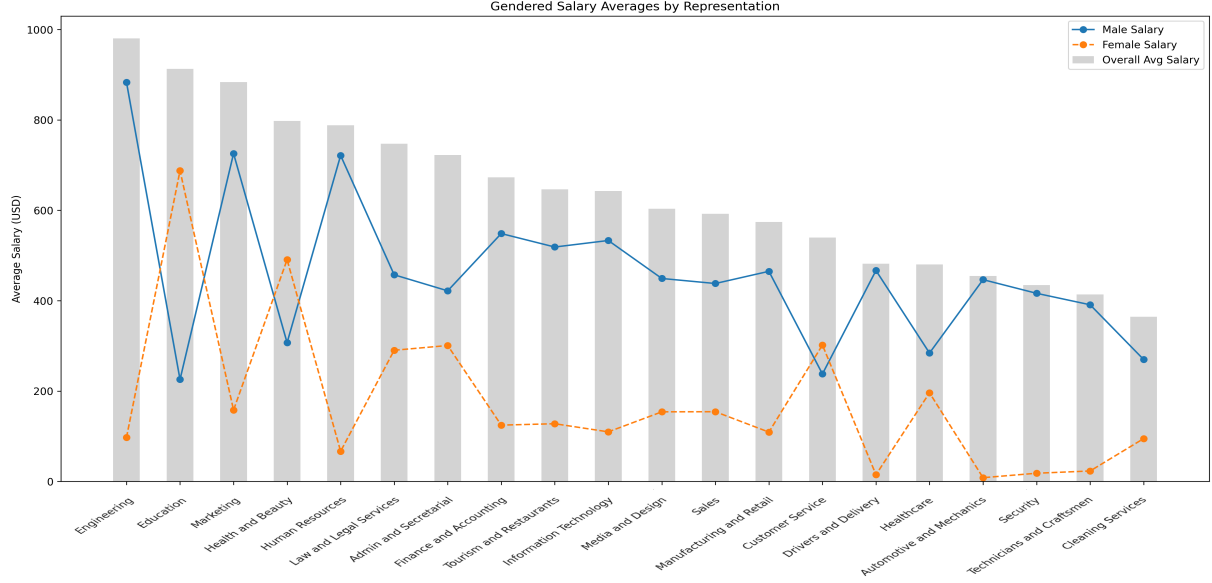

Figure 7: Average salaries in USD by job category, separated by gender and normalised for representation. Bars show overall average; lines indicate male and female-specific averages.

## 7 Linguistic Bias in Arabic Job Ads

To better understand the linguistic framing surrounding gender-targeted language in Arabic job advertisements, we conducted a concordance analysis using a window of ±4 words around selected gendered or appearance-related terms. The analysis was based on tokenisation using the CAMeL Tools Arabic tokenizer for improved segmentation quality (Obeid et al., 2020). Our analysis of Arabic job advertisements reveals a concerning pattern of linguistic bias, especially in job posts targeting women. A range of ads explicitly require candidates to meet criteria unrelated to professional qualifications or experience, focusing instead on appearance, age, and marital status. Table 8 summarises the most frequent patterns we observed.

| Bias Type | Examples from Ads |
|---|---|
| Appearance | جميلة، أنيقة، حسنة المظهر، مظهر لائق، مرتبة، غير محجبة<br>beautiful, elegant, good-looking, decent appearance, tidy, not veiled |
| Personality | لبقة، لبق، لبقة في التعامل، شخصية قيادية، لبقة بالتحدث<br>polite, articulate, good at interaction, leadership personality, well-spoken |
| Age Limits | العمر لا يتجاوز 52 سنة، من 18 إلى 30 سنة، العمر بين 22 و 35<br>age must not exceed 52, from 18 to 30, age between 22 and 35 |
| Marital/Availability Status | عزباء، متفرغة للعمل، غير متزوجة<br>single, available for work, not married |
| Gender Filters | ذكور فقط، إناث فقط، مطلوب شاب، يفضل شابة<br>males only, females only, young man wanted, young woman preferred |
| Emotion/Soft Skill Framing | لبقة، حسنة السلوك، وجه بشوش، حنونة، لبقة مع الزبائن<br>articulate, well-behaved, cheerful face, kind-hearted, good with customers |

Table 8: Examples of Biased Criteria in Arabic Job Advertisements

These include phrases such as جميلة، حسنة المظهر، أنيقة، لبقة، شابة، عزباء، غير محجبة and expressions that specify age limits (e.g., العمر بين 22 و 30) or prefer candidates who are متفرغة (i.e., fully available), sometimes adding that they must be not married (single) غير متزوجة.

Such language reinforces stereotypes about physical attractiveness and gender roles, particularly in roles such as receptionist, sales assistant, or spa worker. Furthermore, certain phrases demand emotional traits like being لبقة (polite/eloquent), which often surface alongside gendered expectations. These requirements, especially when associated with low-skilled roles, suggest systemic patterns of bias and discrimination in hiring.

These phrases indicate structured and recurring forms of discrimination in employment language. A larger sample of concordance examples is included in Appendix B to support transparency and enable further qualitative inspection.

## 8 Conclusion

This paper introduced **ArabJobs**, the first large-scale, publicly available corpus of Arabic job advertisements spanning four Arab countries. The dataset captures linguistic, regional, and socio-economic variation across over 8,500 postings and provides a valuable resource for studying gender representation, dialectal diversity, and occupational language in Arabic. The findings not only validate the quality and versatility of the corpus but also highlight its broader potential to support fairness-aware NLP in under-resourced, real-world contexts. Through a series of experiments, we demonstrated the utility of the dataset for downstream tasks such as salary estimation, job classification, and bias detection. Our analyses revealed systematic gender disparities in both language use and pay, along with clear patterns of occupational segregation. We further showed that large language models like GPT-4 can reliably estimate missing salary information and produce predictions closely aligned with human judgement, reinforcing the value of LLMs in socio-economic text analysis and structured inference.

## Ethical Considerations

The ArabJobs corpus was collected from publicly accessible websites that did not require authentication or payment. Although available in the public domain in practice, the listings are not covered by formal open data licences (e.g., Creative Commons), so the corpus is distributed under a research-only licence for non-commercial academic use. We do not claim ownership of the original content.

All scraping was conducted in compliance with the `robots.txt` directives of the source sites, and no automated access was made to restricted paths. Personally identifiable information was stripped from all records to ensure responsible and ethical data handling.

Table 9 lists the data sources and scraping constraints observed at the time of collection.

| Website | Scraping Allowed? | Notes |
|---|---|---|
| naukrigulf.com | Yes | Avoid listed disallowed paths |
| gulftalent.com | Yes | Do not impersonate blocked bots |
| dubizzle.com | Yes | Avoid disallowed paths, rate-limited |
| tanqeeb.com | Yes | Avoid URLs with parameters |
| jordanrec.com | Yes | Avoid admin/plugin paths |
| forasna.com | Yes | Avoid query filters in URLs |
| sabbar.com | Yes | Fully allowed; provides job sitemaps |

Table 9: Scraping permissions and constraints for the ArabJobs corpus sources.

# Appendix A: Example Prompt for Salary Estimation

Below is a simplified illustration of the structured few-shot prompt used with GPT-4. Three examples with known salaries are provided, followed by one target ad requiring prediction.

```
SYSTEM: You are an assistant that
    predicts monthly salaries
for job ads in Arabic-speaking countries
    .
Always return the salary as <number> <
    currency>.

EXAMPLE 1
Title: Accountant
Location: Cairo, Egypt
Category: Finance and Accounting
Gender: Any
Description: Responsible for financial
    reports and invoices.
Salary: 9,500 EGP

EXAMPLE 2
Title: Sales Executive
Location: Riyadh, Saudi Arabia
Category: Sales
Gender: Male
Description: Outdoor sales for
    electronics company.
Salary: 6,500 SAR

EXAMPLE 3
Title: Nurse
Location: Amman, Jordan
Category: Healthcare
Gender: Female
Description: Provide patient care in
    hospital setting.
Salary: 720 JOD

PREDICT
Title: HR Assistant
Location: Dubai, UAE
Category: Human Resources
Gender: Any
Description: Support recruitment and
    employee records.
Salary:
```

# Appendix B: Job Ads Bias Concordances

Table 10: Biased Criteria in Arabic Job Advertisements - حسنة ، لبقة ، مرتبة، غير محجبة

Table 11: Biased Criteria in Arabic Job Advertisements - متفرغة ، عزباء

Table 12: Biased Criteria in Arabic Job Advertisements - العمر